# Riskyishness and Pinocchio's Search for a Comprehensive Taxonomy of Autonomous Entities

**William Wagner, Anna Żakowska, Clement Aladi, Joseph Santhosh**

Fall 2020, Claremont Graduate University

**Abstract—This paper documents an exploratory pilot study to define the term *Autonomous Entity*, and any characteristics that are required to identify or classify an Autonomous Entity. Our solution builds on previous work with regard to philosophical and scientific classification methods but focuses on a novel Design Science Research Methodology (DSRM) and model to help identify those characteristics which make any autonomous entity similar or different from others. We have solved the problem of not having an existing term to define our lens by creating a new combinatorial term: *Riskyishness*. We present a DSRM and instrument for initial investigation, as well as observational and statistical descriptions of their use in the real world to solicit domain expertise and statistical evidence. Further, we demonstrate a specific application of the methodology by creating a second artifact – a tool to score existing and future technologies based on Riskyishness. The first artifact also provides a technique to disentangle miscellaneous existing technologies or add dimensions to the tools to capture future additions and paradigm shifts.**

## I. INTRODUCTION

Like the Tuscan author Carlo Collodi's wooden puppet, Pinocchio, our project begins the process of answering the question: "What is an Autonomous Entity?" Along the way we ask if there is a way to classify all autonomous entities in the same way that all biological entities can be classified? What are the risks associated with autonomy? And we explore a few other philosophical and logical questions.

Like Pinocchio, we use one experience to define the next. Unlike Pinocchio, however, we are not trying to explore spiritual concepts related to a definition of life. We warrant that there are things a non-biological entity can do that mimic biological behavior exactly, but this does not make them alive in the sense of having a "soul." We have included some of those things in our dimensions from the perspective of observed imitated behavior.

## II. PURPOSE

### A. *AN INTERNAL CHALLENGE - GENERALIZED*

Currently, there is no scientifically accepted definition and/or existing classification of autonomous entities that we could find. We found a plethora of taxonomies that collectively might describe a significant percentage of autonomous entities – none was sufficiently general to meet the needs of our group to find a way to talk about specific features of an autonomous entity from very different perspectives.

Our paper applies an inductive Design Science Research Methodology (DSRM) to the phenetic approach described by McNeil and Heywood as "the production of groups based on overall resemblance shared by their constituent members which permit the largest number of generalizations to be made about them — so called empirical or phenetic taxonomy."[1]

### B. *ADDRESSES MULTIPLE STAKEHOLDERS*

As the world moves into the age of autonomous vehicles, automated assistants for the house and car, and IoT[1] devices that are literally implanted in our vital organs - there is a need for an articulated set of guidelines to help all stakeholders evaluate risks associated with an autonomous entity.

Stakeholders include not only decision-makers and designers, but consumers as well. How can we expect the average consumer to evaluate the risks associated with bringing a given technology into their home if those with advanced understanding cannot agree? Our scoring system's scale can be used with all technologies – from the most benevolent entities, to ones you would definitely not want to bring in your home.

Further, we assert there is a need for a common lexicon of terms to help facilitate a globally applicable taxonomy of autonomous entities viewed through the transdisciplinary lens of Design Science.

### C. *RESEARCH QUESTIONS (RQ)*

- **RQ1**: What is the definition of an "Autonomous Entity"?
- **RQ2**: What are the characteristics that differentiate autonomous entities from each other or other non-autonomous technologies?
- **RQ3**: Is there a common lexicon of terms that can allow transdisciplinary work on complex autonomous entities?
- **RQ4**: Is there a way to quantitatively assess a philosophically sound version of "Risk" with regard to a given entity?

[1] Throughout this paper IoT refers collectively to Internet of Things and Industrial Internet of Things (IIoT). Sometimes referred to as edge devices.

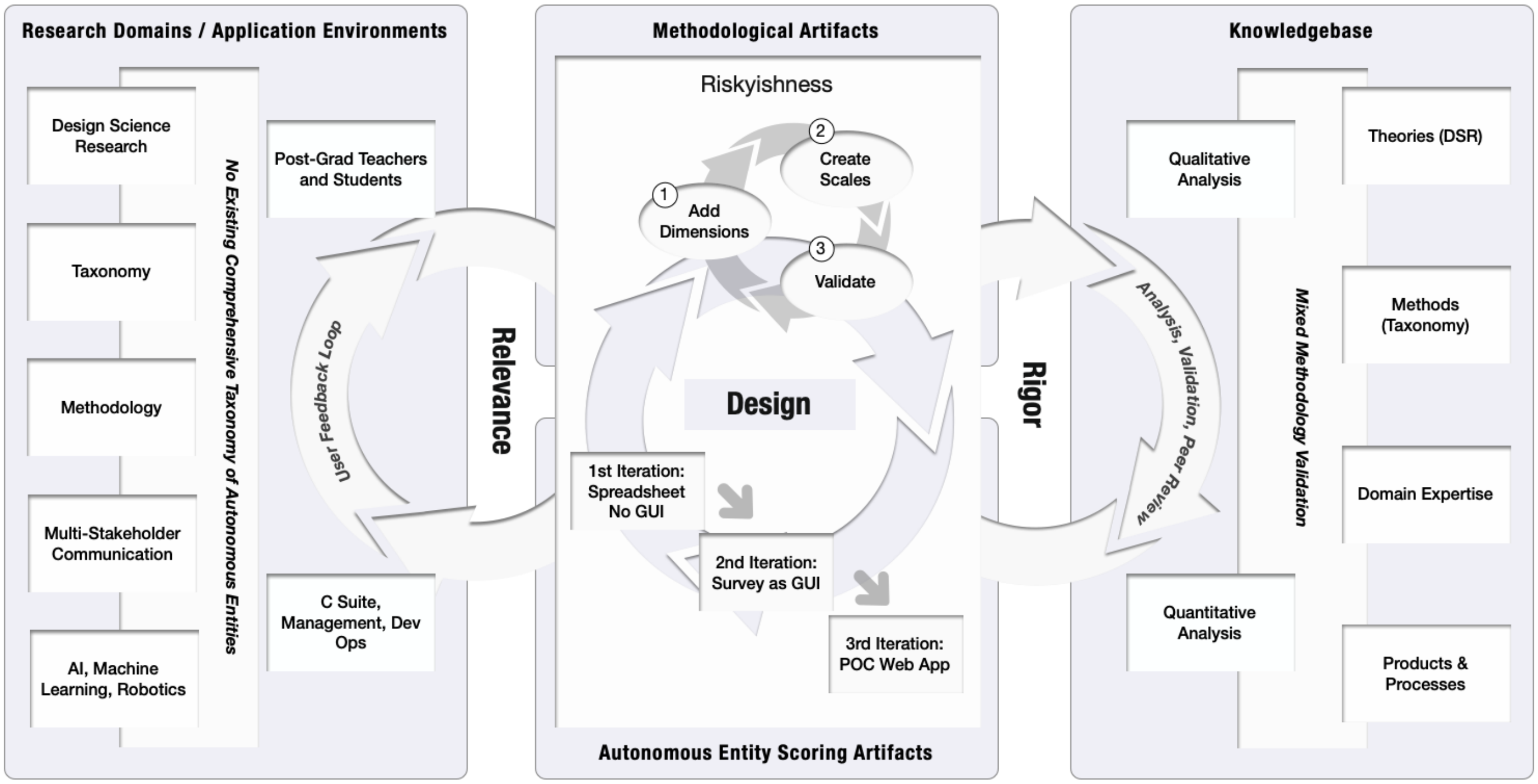

Fig. 1. DSRM Using Riskyishness to Score and Classify Autonomous Entities

Before starting the actual process of articulating a methodology we employed a number of creativity tools to help widen our initial perspectives, then we recursively narrowed the scope and refined the definitions. One of those tools was the Mind Map; developed by Tony Buzan as a tool to help groups be more creative in their thinking[2]. We found it helped us to articulate a variety of concerns with regard to defining an autonomous entity.

Our original list of dimensions only had 12 dimensions from a brainstorming session and the mind map helped us go from 12 to 25, as well as adding in the Classes, which our initial group did not have. Then we had something to iterate upon per the DSRM.

A second benefit of the mind map was that it exposed an initial weakness in our approach – that of making the idea of AI central to the taxonomy. This shift in thinking helped us to adjust our taxonomy to include AI via indirect constructs, thus making it far more generalizable.

## III. Theoretical Framework

### A. Design Science Research Methodology

Our solution builds on the work of Hevner, Chatterjee, Peffers, and others in that we draw heavily on the Design Science Research Methodology (DSRM).[3][4] DSRM provides the requisite framework to merge real-world concerns of artifact design and creation with the philosophical and academic concerns of the ethics associated with technology.

Our **H1** hypothesis is that there is a scientifically sound way to create a representative construct capable of measuring a wide variety of physical and philosophical dimensions across disparate technologies.

We have solved the problem of not having an existing term or endogenous[5] construct to define our lens by creating a new term: "Riskyishness."

Our model (Fig.1) employs the cyclical nature of DSR to refine our philosophical and scientific classification methods but focuses on a novel methodology and framework to help identify those characteristics which make any autonomous entity similar or different from others.

Our project meets the three foundational requirements of DSR by:

1 - Demonstrating **relevance** in the application space. This is supported by the responses (n~30) from an initial survey asking about the relevance of 26 dimensions with regard to classifying an entity.

2 - Two distinct **design artifacts**.

A) A design methodology and instrument for initial investigation. We used our own techniques to design the techniques for design – hopefully validating both in the process.

B) A real-world Proof-of-Concept (POC) web app to allow users to score any entity for riskyishness. We iterated through three versions of a tool to allow users to score any entity based on the dimensions defined by the first artifact.

3 - Finally, our DSR model maintains its philosophical **rigor** and contributes at a theoretical level by demonstrating a novel technique for classification supported by a mixed-method validation process employing both Qualitative and Quantitative analysis.

### B. Taxonomy and Classification

#### 1) Kernel Theory

In 1964 McNeil et al. described a rising methodology known as the "numerical" approach. Perhaps prescient at the time, the "numerical" goal was, "to eliminate much of the subjectivity of the processes of classification by endeavoring to specify all the attributes of the organisms and then making a mathematical assessment of similarity, the scale of the calculations requiring the use of electronic computers."[1]

This "numerical" approach (today it might be called Big Data[6], Machine Learning[7], or AI[8]) greatly favored the phenetic over the evolutionary approach to taxonomy in that it made classifications that were not based on evolutionary reasons – a central tenet and reciprocal *raison d'être* of biological taxonomy.

Traditional biological taxonomies could still be represented if one chose to make biological evolution the reason for a classification – this is known as cladistics[9]. But not only is the biological lineage (or development lineage if classifying technology) no longer the universal reason to split a group, it now is not even necessary at all. One can now represent groups whose lifecycles may not always develop linearly. Further, arbitrary classification is also made possible, provided there is an underlying logical framework that is universally applied for classification.

#### 2) Modern Taxonomies

In looking at both classic and modern taxonomies we found an overriding characteristic of every taxonomy was that they were born of necessity. In every case - including our own and the classic taxonomies - the taxonomy was a result of a group's specific need to classify entities in order to facilitate **other research** or **other activities** like purchase decisions.

However, while McNeil and others cited in the kernel theory demonstrate generalizability and a logical framework, we found that many of the modern examples misuse the term taxonomy to include what is really just a well-researched feature comparison.

We have included three representative examples of taxonomies we found which at least satisfy the requirement of convincing us that one could use their taxonomy for their stated purpose. We found others that we have not included that failed to convince us that they had sufficient domain expertise to justify their taxonomy's ability to make even limited, non-generalizable, specific recommendations.

The following three taxonomies demonstrate good domain knowledge, and we would accept their respective taxonomies as authoritative within a very narrow scope. It was the philosophical and logical foundations that we found lacking in all but the most rigorous taxonomies.

*a) Successful Taxonomy, Successful Foundations, Too Narrow a Scope to be Applicable to Our Needs*

Arya et al. describe in great detail a "taxonomy of toolkits for AI explainability."[10] Their specific use case is not generalizable beyond the software they are classifying because of the nature of their taxonomy being derivative of the inner mathematics in AI.

They do, however, demonstrate a valid use of taxonomy to classify complex technology, demonstrating good philosophical and mathematical foundations that allow the reader to accept their descriptions of constructs that by their very nature try to explain what might otherwise be unexplainable.

*b) Successful Taxonomy, Some Foundational Support, Too Narrow a Scope to be Applicable to Our Needs*

Cecile Yehezkel's "Taxonomy of Computer Architecture Visualizations" attempts to straddle the boundaries of hardware and software at the lowest levels with a "taxonomy focusing on [computer architecture and assembly language] with emphasis on the didactic and cognitive aspect of the visualization environment."[11] Unfortunately, Yehezkel demonstrates an incomplete taxonomy in that it failed to provide a sufficient basis for several of the generalizations in the paper.

The taxonomy appears valid for the extremely narrow scope of its intended audience - instructors evaluating assembly language visualization software – but without more supporting evidence it is little more than a fancy software comparison. It has limited generalizability. The author does not try to hide this fact, which lends credibility to the taxonomy.

*c) Successful Taxonomy, Too Little Foundational Support and Perhaps Incorrectly Used, Attempted Scope is Fine but Methodology Fails to Generalize to Match the Scope*

Dombroviak and Ramnath propose a "Taxonomy of Mobile and Pervasive Technologies."[12] Again, while some of the dimensions they have extracted may be interesting to think about, their N of only 14 "sample packages from a range of domains" leaves much to be desired, especially since there is no other validation provided. Also, there is no foundational theory provided that explains how one dimension is compared to another.

They suggest that the logic diagram and explanation equate to something approaching statistical separation. This is not the case. They defined the categories based on already-selected software; one can almost always define groups in a post-hoc way that would meet their rules for logic. This is a classic example of over-fitting. Therefore, we agree that while those rules might be a requirement for a successful taxonomy, in and of themselves they do not constitute sufficient philosophical or logical support for generalization. In summary, the same information could be presented in a single software feature comparison chart.

## IV. The Artifacts

### A. Methodology

Our project demonstrates the application and use of a modern DSR methodology for classifying autonomous entities. This method looks at technology through the lens of "Riskyishness." Once applied, this concept can be used to score any piece of technology; as well as provide a practical means of evaluating areas of concern for multiple stakeholders and effectively sharing those concerns both horizontally and vertically across an organization or discipline.

The artifact articulates our definition for two key terms - Autonomous Entity and Riskyishness.

*1) Autonomous Entity*

Our methodology defines an Autonomous Entity as:

**"Any type of technology that interacts with its environment and reacts based on environmental stimuli."**

In other words, no AI or even digital capability is required for a technology to be autonomous. It is very easy to imagine a robot capable of autonomously searching the edges of a room for an opening using only analog sensors and actuators.

However, our definition does differentiate autonomous technology from non-autonomous technology like the aforementioned analog robot vs. an old combustion engine, which would not be considered autonomous; however, as we will see, the engine could still be evaluated using Riskyishness.

We want our definition to be all-inclusive of autonomy with regard to technology. Here is a sample list of Autonomous Entities in no particular order:

- Siri™
- Alexa™
- An assembly line robot
- A robotic dog
- An emergency response drone
- A smart refrigerator
- Roomba™
- Apple Watch™
- A computer virus
- Raspberry Pi™
- Simulation agents
- YouTube™ Recommendation Engine
- PageRank™ Algorithm

*2) Riskyishness*

We have created an endogenous construct called Riskyishness. We used a combinatory term like "Risky-ish-ness" to provide a guide to creating the scales. *We do not try to articulate specific risks associated with a given dimension - instead, riskyishness allows us to make generalizations.*

For example, we warrant that any device that is networked has inherently more Riskyishness than a standalone version of the same thing. We do not seek to articulate the specific ways that a networked entity can attack or be attacked, only acknowledge that networking provides additional risks.

This allows us to use the same 5-point scale (0 to 4) to compare disparate dimensions like Size and Data Storage.

Typically, a measurement of physical size would not be an appropriate measure to also describe data storage, and vice versa. Imagine describing the length of a very big truck in terms of gigabytes, or hard drive storage as 51 feet. However we can say that an autonomous truck inherently has more Riskyishness associated with it than a bicycle, and the truck is a 4 on the scale of Riskyishness.

And we can say that remote, long-term data storage has more inherent Riskyishness than the same device that does not store any data. Remote data storage is also a 4 on the scale of Riskyishness.

Both the Data Storage and the Size can be said to be 4 Riskyishness units in magnitude - allowing us to use the same idea to compare disparate things.

Our methodology proposes 25 dimensions grouped into six classes (Table 2) that we believe are collectively capable of identifying different types of technology and provides a means to categorize all known and future autonomous entities. We also created a Lexicon of Terms used in our approach (Appendix A)

While the end-user will see a predefined list of dimensions, our research process requires us to define the lens used to create those dimensions, as well as the scale used to measure each dimension (Appendix B).

Some entities may not display all of the dimensions described, just like all animals do not have gills or wings; however, we believe the methodology to be generalizable, with an eye towards future paradigm shifts like quantum computing-based neural nets capable of completely unsupervised learning on unstructured data that build their own hardware and write their own software.

To help establish the 25 dimensions, six classes, and the validity of the scales used to measure each one, we created two Qualtrics surveys. Initially, we created a single survey to capture initial impressions of our dimensions as well as collecting some sample entities to test the methodology. After soliciting preliminary editorial and survey design feedback (more than 125 collective years of research experience) from multiple CGU faculty, outside industry, and others, we divided the survey into two instruments and focused on making them faster and shorter.

The first instrument is designed to solicit domain expertise with regard to the dimensions and the scales. The goal of this survey is to help define the construct of Riskyishness without specifically asking the participants to think philosophically. We hope that by providing the list of dimensions we have created and well-defined scales, the initial domain expertise can be aggregated, coded, and used to identify those dimensions that experts have found to be particularly useful.

We also provided an open-ended question to ask for anything we might have missed. We acknowledge the sample size ($n_{Framework}$) of respondents at the beginning to be statistically less than optimal for a more substantial investigation, however, we hope that the collective domain expertise still rises to an acceptable standard for the initial construction of the framework. We chose a semi-arbitrary goal of N=30 and got N=29 respondents (28 valid) drawn from CGU professors, students, industry contacts, and other associates. There were an additional 30 surveys that were less than 30% completed.

We intentionally included a range of respondents because as stated earlier, while the complex concepts should be formulated and articulated with domain expertise, the ideas should be easily communicable to anyone.

We did not collect any personally identifiable information or any kind of digital fingerprints. Each response was given a unique id by the survey system like: R_1jTzXCKz9JtKB7T. In theory it would have been possible for someone to take the survey twice, but because this is a pilot, and we had enough difficulty getting people to take it the first time, we decided the increase in people's willingness to A) take the time and B) be honest, outweighed any need to collect more info and drive people away. We are confident none of the responses are duplicated but must acknowledge it could be a possibility.

Before asking any questions, the survey starts by providing the respondents with an introductory description of Riskyishness and the same definition of *autonomous entity* presented in this paper.

*a) 3 Ethnographic Questions*

TABLE 1 – ETHNOGRAPHIC PREVIOUS EXPERIENCE RESULTS

| | Robotics, AI, or Machine Learning | Autonomous Entities | Professional Experience |
|---|---|---|---|
| ***count*** | 28 | 28 | 28 |
| ***mean*** | 2.071428571 | 2.178571429 | 2.321428571 |
| ***skew*** | 0.290030894 | 0.120590576 | -0.58436131 |
| ***kurtosis*** | 2.151104196 | 0.261059747 | -0.810460178 |
| ***std*** | 0.465758754 | 0.547964005 | 0.722832465 |
| ***min*** | 1 | 1 | 1 |
| ***25%*** | 2 | 2 | 2 |
| ***50%*** | 2 | 2 | 2 |
| ***75%*** | 2 | 2.25 | 3 |
| ***max*** | 3 | 3 | 3 |

The first section asks three ethnographic questions using a simple 3-point scale. The goal of the section was to get a brief measure of our respondents' familiarity with the domain of robotics, AI, and machine learning, as well as a gauge of their understanding of autonomous entities as we had described.

The distribution of responses showed the range of input for which we had hoped, with a slight negative skew to the right, indicating that we had more expert or advanced knowledge than beginners.

On the first question, asking if the respondents knew about robotics, AI, or machine learning, 22 respondents (78.57%) identified themselves in the middle group, and four (14.29%) self-identified as experts. Two responded that they "Didn't know about Robots, AI, or Machine Learning."

The second question, which asked about autonomous entities, showed 7 respondents (25%) self-identified as having a good idea of what an autonomous entity might be, and 19 people (67%) said they have some ideas about what an autonomous entity might be.

The third question asked about professional experience. Almost half of the respondents, 13 people (46.43%), work or worked in a computer, robotics, machine learning, or a related scientific field, 11 people (39.29%) said they used computers but don't know how they work, and the final 4 people (14.29%) said their work had nothing to do with computers or robots.

*b) Class Questions*

For the remainder of the survey, other than the final open-ended text-based response, we used a 4-point Likert scale. We used an even-numbered point scale to remove the possibility of neutral responses.

For each Class we asked two questions. The Class questions were presented with individual 4-point scales for each question. The respondents were shown a table of the Class (Fig.2).

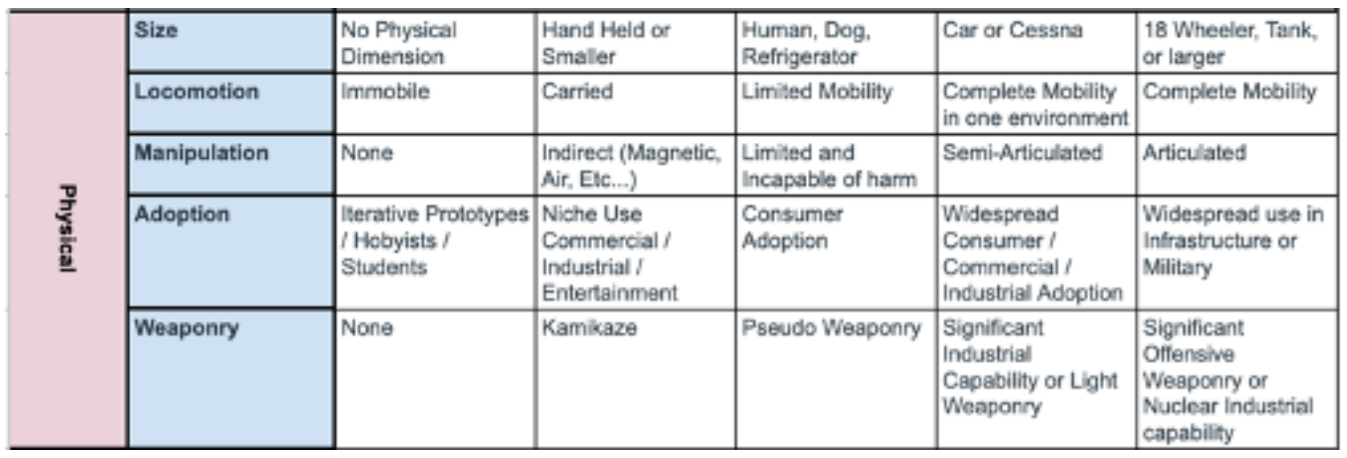

| Physical | | | | | | |
|---|---|---|---|---|---|---|
| | **Size** | No Physical Dimension | Hand Held or Smaller | Human, Dog, Refrigerator | Car or Cessna | 18 Wheeler, Tank, or larger |
| | **Locomotion** | Immobile | Carried | Limited Mobility | Complete Mobility in one environment | Complete Mobility |
| | **Manipulation** | None | Indirect (Magnetic, Air, Etc...) | Limited and Incapable of harm | Semi-Articulated | Articulated |
| | **Adoption** | Iterative Prototypes / Hobyists / Students | Niche Use Commercial / Industrial / Entertainment | Consumer Adoption | Widespread Consumer / Commercial / Industrial Adoption | Widespread use in Infrastructure or Military |
| | **Weaponry** | None | Kamikaze | Pseudo Weaponry | Significant Industrial Capability or Light Weaponry | Significant Offensive Weaponry or Nuclear Industrial capability |

Fig. 2. Physical Class

TABLE 2 – CLASSES AND DIMENSIONS

| Physical | UI/UX | Ethical / Philosophical | Application | Security/Privacy | Embodiment |
|---|---|---|---|---|---|
| Size | Human Interaction | Local Comprehension | Intended Use | Susceptible to Outside Influence | Physicality |
| Locomotion | # of Human Senses addressed by UI/UX | Global Comprehension | Financial Resources | 3rd Party Dependencies | Experiential |
| Manipulation | | Decision Making | | Data Collection | Emotional |
| Adoption | | Contextualization | | Data Storage | Social |
| Weaponry | | Self-Preservation | | Data Usage | Transcendental |
| | | Complexity | | | |
| Size | No Physical Dimension | Hand Held or Smaller | Human, Dog, Refrigerator | Car or Cessna | 18 Wheeler, Tank, or larger |

Fig. 3. Size Dimension

An interesting insight from the responses showed that there were a number of comments that indicated that respondents often felt that a given class of dimensions was necessary for classification, however they often disagreed with the specific word we had used to describe or name the class. Our quantitative analysis of the responses showed that in general, respondents approved of the classes.

*c) Dimension Questions*

For efficiency, the Dimensions were presented as a matrix, but they still used a 4-point Likert scale. This made the survey much faster to answer, as well as reducing the number of questions. Each Dimension's row (Fig.3) from the table was shown for each matrix.

Our goal for the questions focused on the dimensions was both to gauge the effectiveness of the scales in communicating a given dimension, as well as to determine if the scale actually measured the dimension.

The responses once again indicate that overall, respondents were satisfied with the dimensions. The mean value for the necessity of all dimensions was 2.78 on a 4-point scale. It is perhaps interesting to note that security and privacy issues appear to be considered the most important for classifying an entity, with 4 out of 5 of the most important dimensions as identified be the respondents. The two highest ranked dimensions both had a mean of 3.07, placing them significantly higher than the lowest which was Weaponry, having a mean score of only 2.28.

Users also found that Weaponry had the least descriptive scale as well as the lowest score with regards to how well the scale measures the range of possibilities.

*a) Final Question*

We concluded the survey with an open question. In reviewing the comments, it is clear that although we tried to make the scale accessible to a wide range of users, there is still significant room for improvement regarding the clarity of some of the technical terms. In other words, some of the concepts are so technical that even our attempt at a simplified non-technical description was still overly dependent on technical language.

Respondents said the lexicon greatly aided in their understanding of a given dimension, however the goal was to create a stand-alone tool that was sufficiently intuitive that it could be used without the lexicon.

This will be an area of further study. The specific mechanics and methodolgy of creating a scale for a given dimension merits detailed examination. We believe it may be possible to integrate upvoting a given scale with a public version of the tool discussed in the next section to create a mechanism by which the scales can be evaluated and improved by the users. This same mechanism might also work to create self-regulated weights for the dimensions.

### B. *Autonomous Entity Scoring Tool*

A second artifact, distinct from the framework, is a tool to demonstrate the framework's use and allow the user to customize the methodology to match their domain or application. The website provides further user experience by allowing the user to prioritize aspects of the methodology for their specific application.

*1) First Iteration*

The first iteration of the tool was a simple spreadsheet. One had to look up the scale on a separate table and enter the appropriate number in the spreadsheet. It grew from 15 columns, to 20, then finally to 25 dimensions. The spreadsheet also calculates a basic score based on the mean of the answers.

We did a spot check at multiple times throughout the process using 10 miscellaneous autonomous entities chosen to represent different types. This informal testing allowed us to refine the process and think about possible Graphical User Interfaces (GUIs) and formulas for scoring an entity.

*2) Second Iteration*

The second iteration of the tool uses a survey as a GUI to collect examples of various technologies. The survey is the second half of the long survey originally created that we split in half. This allowed us to reuse an element and saved significant development time. It uses hot spots on an image of the table row to allow the user to see the description and click, rather than having to look up the info and type it in.

Having a GUI cut the time to enter an entity in half, from about 7-10 minutes to 2-5 minutes. It also reduced errors. We have not tracked specifically the number or magnitude of error, however we noticed when re-entering the entities from the first tool into the second that there were clear mistakes, most likely attributable to having to look up the scale before entering it on the spreadsheet. That analysis can be done post-hoc and will likely be included in further papers. We have accumulated a sample (n) of 86 entities. Again, these answers are coded to allow for analysis in aggregate.

The entities from the tool are scored to provide dual output. One output is a sample contribution to the pool of entities available to the cluster (numerical) analysis of the responses; the second output is a score that allows one to evaluate the riskyishness of a given entity. In initial spot testing on a variety of entities, we have found the scale to accurately measure general riskyishness between very different types of technologies. It works in both relative and absolute terms. In other words, riskier Entities relative to each other score higher, and the absolute comparisons of those Entities with a higher Score show they are riskier than those with lower.

TABLE 4 – SURVEY 1 – CLASS DESCRIPTION SUMMARY STATISTICS

| | Physical Describe? | Physical Necessary? | UI/UX Describe? | UI/UX Necessary? | Ethical / Philosophical Describe? | Ethical / Philosophical Necessary? | Application Describe? | Application Necessary? | Security / Privacy Describe? | Security / Privacy Necessary? | Embodiment Describe? | Embodiment Necessary? |
|---|---|---|---|---|---|---|---|---|---|---|---|---|
| **count** | 28 | 28 | 27 | 27 | 28 | 28 | 28 | 28 | 28 | 28 | 28 | 28 |
| **mean** | 2.53571429 | 3.10714286 | 2.48148148 | 2.7037037 | 2.32142857 | 2.82142857 | 2.64285714 | 2.96428571 | 2.78571429 | 3.21428571 | 2.64285714 | 2.82142857 |
| **median** | 2.5 | 3 | 2 | 2 | 2 | 3 | 2.5 | 3 | 3 | 3 | 3 | 3 |
| **skew** | 0.02575042 | -0.6897157 | 0.15982826 | 0.14787785 | 0.41183466 | -0.6231438 | 0.06556236 | -0.4149417 | -0.1614036 | -0.850605 | -0.3051357 | -0.2690395 |
| **kurtosis** | -0.8492253 | 0.44822051 | -1.1210065 | -1.2643693 | -0.838062 | -0.4267872 | -1.0663606 | -0.0677252 | -0.8145817 | 0.23083969 | -0.9448293 | -0.6611575 |
| **std** | 0.96156288 | 0.78595475 | 1.05138628 | 0.99285195 | 1.02029667 | 0.98332661 | 0.9893614 | 0.79265811 | 0.91720763 | 0.8325393 | 1.02611405 | 0.90486633 |
| **min** | 1 | 1 | 1 | 1 | 1 | 1 | 1 | 1 | 1 | 1 | 1 | 1 |
| **25%** | 2 | 3 | 2 | 2 | 2 | 2 | 2 | 2.75 | 2 | 3 | 2 | 2 |
| **50%** | 2.5 | 3 | 2 | 2 | 2 | 3 | 2.5 | 3 | 3 | 3 | 3 | 3 |
| **75%** | 3 | 4 | 3 | 4 | 3 | 3.25 | 3.25 | 3.25 | 3.25 | 4 | 3 | 3.25 |
| **max** | 4 | 4 | 4 | 4 | 4 | 4 | 4 | 4 | 4 | 4 | 4 | 4 |

TABLE 5 – DIMENSION – NECESSARY?

| Questions | count | mean | std | min | 25% | 50% | 75% | max |
|---|---|---|---|---|---|---|---|---|
| **Data Collection: Necessary?** | 27 | 3.07407407 | 0.78082431 | 1 | 3 | 3 | 4 | 4 |
| **Self-Preservation: Necessary?** | 28 | 3.07142857 | 0.93999887 | 1 | 2.75 | 3 | 4 | 4 |
| **Data Storage: Necessary?** | 27 | 3.03703704 | 0.80772622 | 1 | 3 | 3 | 4 | 4 |
| **Susceptible to Outside Influence: Necessary?** | 28 | 3.03571429 | 0.83808171 | 1 | 2.75 | 3 | 4 | 4 |
| **Locomotion: Necessary?** | 28 | 2.92857143 | 0.89973541 | 1 | 2 | 3 | 4 | 4 |
| **Human Interaction: Necessary?** | 28 | 2.89285714 | 0.9940298 | 1 | 2 | 3 | 4 | 4 |
| **Intended Use: Necessary?** | 28 | 2.89285714 | 0.87514171 | 1 | 2 | 3 | 4 | 4 |
| **Transcendental: Necessary?** | 27 | 2.88888889 | 1.05003052 | 1 | 2 | 3 | 4 | 4 |
| **Social: Necessary?** | 26 | 2.88461538 | 0.99305279 | 1 | 2 | 3 | 4 | 4 |
| **Adoption: Necessary?** | 28 | 2.82142857 | 0.86296523 | 1 | 2 | 3 | 3 | 4 |
| **Complexity: Necessary?** | 28 | 2.82142857 | 1.02029667 | 1 | 2 | 3 | 4 | 4 |
| **Decision Making: Necessary?** | 28 | 2.82142857 | 0.86296523 | 1 | 2 | 3 | 3.25 | 4 |
| **3rd Party Dependencies: Necessary?** | 27 | 2.81481481 | 0.83376057 | 1 | 2 | 3 | 3 | 4 |
| **# of Human Senses: Necessary?** | 27 | 2.74074074 | 0.90267093 | 1 | 2 | 3 | 3 | 4 |
| **Data Usage: Necessary?** | 27 | 2.74074074 | 0.85900625 | 1 | 2 | 3 | 3 | 4 |
| **Physicality: Necessary?** | 27 | 2.74074074 | 0.94431874 | 1 | 2 | 3 | 3 | 4 |
| **Financial Resources: Necessary?** | 27 | 2.7037037 | 0.99285195 | 1 | 2 | 3 | 3.5 | 4 |
| **Emotional: Necessary?** | 26 | 2.69230769 | 1.04954202 | 1 | 2 | 3 | 3.75 | 4 |
| **Local Comprehension Method: Necessary?** | 28 | 2.67857143 | 0.90486633 | 1 | 2 | 3 | 3 | 4 |
| **Contextualization: Necessary?** | 28 | 2.64285714 | 1.02611405 | 1 | 2 | 3 | 3.25 | 4 |
| **Size: Necessary?** | 28 | 2.64285714 | 0.95118973 | 1 | 2 | 2 | 3.25 | 4 |
| **Manipulation: Necessary?** | 28 | 2.60714286 | 0.95604454 | 1 | 2 | 2.5 | 3 | 4 |
| **Experiential: Necessary?** | 27 | 2.59259259 | 1.04731375 | 1 | 2 | 2 | 3.5 | 4 |
| **Global Comprehension Method: Necessary?** | 27 | 2.51851852 | 0.93522397 | 1 | 2 | 2 | 3 | 4 |
| **Weaponry: Necessary?** | 28 | 2.28571429 | 1.08379111 | 1 | 1 | 2 | 3 | 4 |

TABLE 6 – DIMENSION – GOOD SCALES?

| Questions | count | mean | std | min | 25% | 50% | 75% | max |
|---|---|---|---|---|---|---|---|---|
| **Data Collection: Good Scale?** | 27 | 3.11111111 | 0.80064077 | 1 | 3 | 3 | 4 | 4 |
| **Data Storage: Good Scale?** | 27 | 3.11111111 | 0.89155583 | 1 | 3 | 3 | 4 | 4 |
| **Locomotion: Good Scale?** | 28 | 3.03571429 | 0.83808171 | 1 | 2.75 | 3 | 4 | 4 |
| **Self-Preservation: Good Scale?** | 28 | 3.03571429 | 0.99933841 | 1 | 2 | 3 | 4 | 4 |
| **Human Interaction: Good Scale?** | 28 | 2.96428571 | 0.92224135 | 1 | 2 | 3 | 4 | 4 |
| **Transcendental: Good Scale?** | 27 | 2.96296296 | 0.93978236 | 1 | 3 | 3 | 4 | 4 |
| **Emotional: Good Scale?** | 26 | 2.92307692 | 1.01678225 | 1 | 2 | 3 | 4 | 4 |
| **Social: Good Scale?** | 26 | 2.88461538 | 0.90893006 | 1 | 2 | 3 | 4 | 4 |
| **Complexity: Good Scale?** | 28 | 2.82142857 | 1.05597318 | 1 | 2 | 3 | 4 | 4 |
| **Intended Use: Good Scale?** | 28 | 2.82142857 | 0.90486633 | 1 | 2 | 3 | 4 | 4 |
| **Susceptible to Outside Influence: Good Scale?** | 28 | 2.82142857 | 0.98332661 | 1 | 2 | 3 | 4 | 4 |
| **Adoption: Good Scale?** | 28 | 2.78571429 | 0.91720763 | 1 | 2 | 3 | 3.25 | 4 |
| **Decision Making: Good Scale?** | 28 | 2.75 | 0.79930525 | 1 | 2 | 3 | 3 | 4 |
| **Physicality: Good Scale?** | 27 | 2.74074074 | 0.98420576 | 1 | 2 | 3 | 3.5 | 4 |
| **# of Human Senses: Good Scale?** | 27 | 2.7037037 | 0.86889916 | 1 | 2 | 3 | 3 | 4 |
| **3rd Party Dependencies: Good Scale?** | 27 | 2.7037037 | 0.82344561 | 1 | 2 | 3 | 3 | 4 |
| **Size: Good Scale?** | 28 | 2.64285714 | 0.95118973 | 1 | 2 | 2 | 3.25 | 4 |
| **Data Usage: Good Scale?** | 27 | 2.62962963 | 0.88353086 | 1 | 2 | 3 | 3 | 4 |
| **Financial Resources: Good Scale?** | 27 | 2.59259259 | 1.00992228 | 1 | 2 | 3 | 3 | 4 |
| **Local Comprehension Method: Good Scale?** | 28 | 2.57142857 | 0.79015101 | 1 | 2 | 3 | 3 | 4 |
| **Manipulation: Good Scale?** | 28 | 2.53571429 | 0.96156288 | 1 | 2 | 2 | 3 | 4 |
| **Contextualization: Good Scale?** | 27 | 2.51851852 | 1.05138628 | 1 | 2 | 2 | 3 | 4 |
| **Experiential: Good Scale?** | 27 | 2.44444444 | 1.15470054 | 1 | 1 | 3 | 3 | 4 |
| **Global Comprehension Method: Good Scale?** | 27 | 2.37037037 | 0.9666814 | 1 | 2 | 2 | 3 | 4 |
| **Weaponry: Good Scale?** | 28 | 2.35714286 | 0.91142078 | 1 | 2 | 2 | 3 | 4 |

TABLE 7 – DIMENSION – HELPFUL DESCRIPTIONS?

| Questions | count | mean | std | min | 25% | 50% | 75% | max |
|---|---|---|---|---|---|---|---|---|
| **Data Storage: Helpful Desc?** | 27 | 3.18518519 | 0.92141351 | 1 | 3 | 3 | 4 | 4 |
| **Data Collection: Helpful Desc?** | 27 | 3.14814815 | 0.81823937 | 1 | 3 | 3 | 4 | 4 |
| **Locomotion: Helpful Desc?** | 28 | 3.07142857 | 0.81325006 | 1 | 3 | 3 | 4 | 4 |
| **Human Interaction: Helpful Desc?** | 28 | 3.03571429 | 0.83808171 | 1 | 2.75 | 3 | 4 | 4 |
| **Emotional: Helpful Desc?** | 26 | 3 | 0.93808315 | 1 | 2.25 | 3 | 4 | 4 |
| **Self-Preservation: Helpful Desc?** | 28 | 3 | 0.98130676 | 1 | 2 | 3 | 4 | 4 |
| **Social: Helpful Desc?** | 26 | 3 | 0.89442719 | 1 | 2 | 3 | 4 | 4 |
| **Adoption: Helpful Desc?** | 28 | 2.92857143 | 0.97860715 | 1 | 2 | 3 | 4 | 4 |
| **Susceptible to Outside Influence: Helpful Desc?** | 28 | 2.92857143 | 0.85758366 | 1 | 2 | 3 | 4 | 4 |
| **Transcendental: Helpful Desc?** | 27 | 2.92592593 | 0.99714693 | 1 | 2 | 3 | 4 | 4 |
| **Data Usage: Helpful Desc?** | 27 | 2.85185185 | 0.90739287 | 1 | 2 | 3 | 4 | 4 |
| **Intended Use: Helpful Desc?** | 28 | 2.82142857 | 0.90486633 | 1 | 2 | 3 | 3.25 | 4 |
| **Size: Helpful Desc?** | 28 | 2.82142857 | 0.81892302 | 2 | 2 | 3 | 3.25 | 4 |
| **Decision Making: Helpful Desc?** | 28 | 2.78571429 | 0.8325393 | 1 | 2 | 3 | 3 | 4 |
| **3rd Party Dependencies: Helpful Desc?** | 27 | 2.77777778 | 0.84731855 | 1 | 2 | 3 | 3 | 4 |
| **Physicality: Helpful Desc?** | 27 | 2.77777778 | 0.97402153 | 1 | 2 | 3 | 3.5 | 4 |
| **Complexity: Helpful Desc?** | 28 | 2.75 | 0.92796073 | 1 | 2 | 3 | 3.25 | 4 |
| **Financial Resources: Helpful Desc?** | 27 | 2.59259259 | 0.97109214 | 1 | 2 | 2 | 3 | 4 |
| **Manipulation: Helpful Desc?** | 28 | 2.57142857 | 0.99735099 | 1 | 2 | 2.5 | 3 | 4 |
| **# of Human Senses: Helpful Desc?** | 27 | 2.55555556 | 0.93369956 | 1 | 2 | 2 | 3 | 4 |
| **Contextualization: Helpful Desc?** | 28 | 2.53571429 | 1.07089934 | 1 | 2 | 2 | 3.25 | 4 |
| **Local Comprehension Method: Helpful Desc?** | 28 | 2.53571429 | 0.88116686 | 1 | 2 | 2 | 3 | 4 |
| **Experiential: Helpful Desc?** | 27 | 2.51851852 | 1.22066723 | 1 | 1 | 3 | 4 | 4 |
| **Global Comprehension Method: Helpful Desc?** | 28 | 2.46428571 | 0.92224135 | 1 | 2 | 2 | 3 | 4 |
| **Weaponry: Helpful Desc?** | 28 | 2.46428571 | 1.07089934 | 1 | 2 | 2 | 3 | 4 |

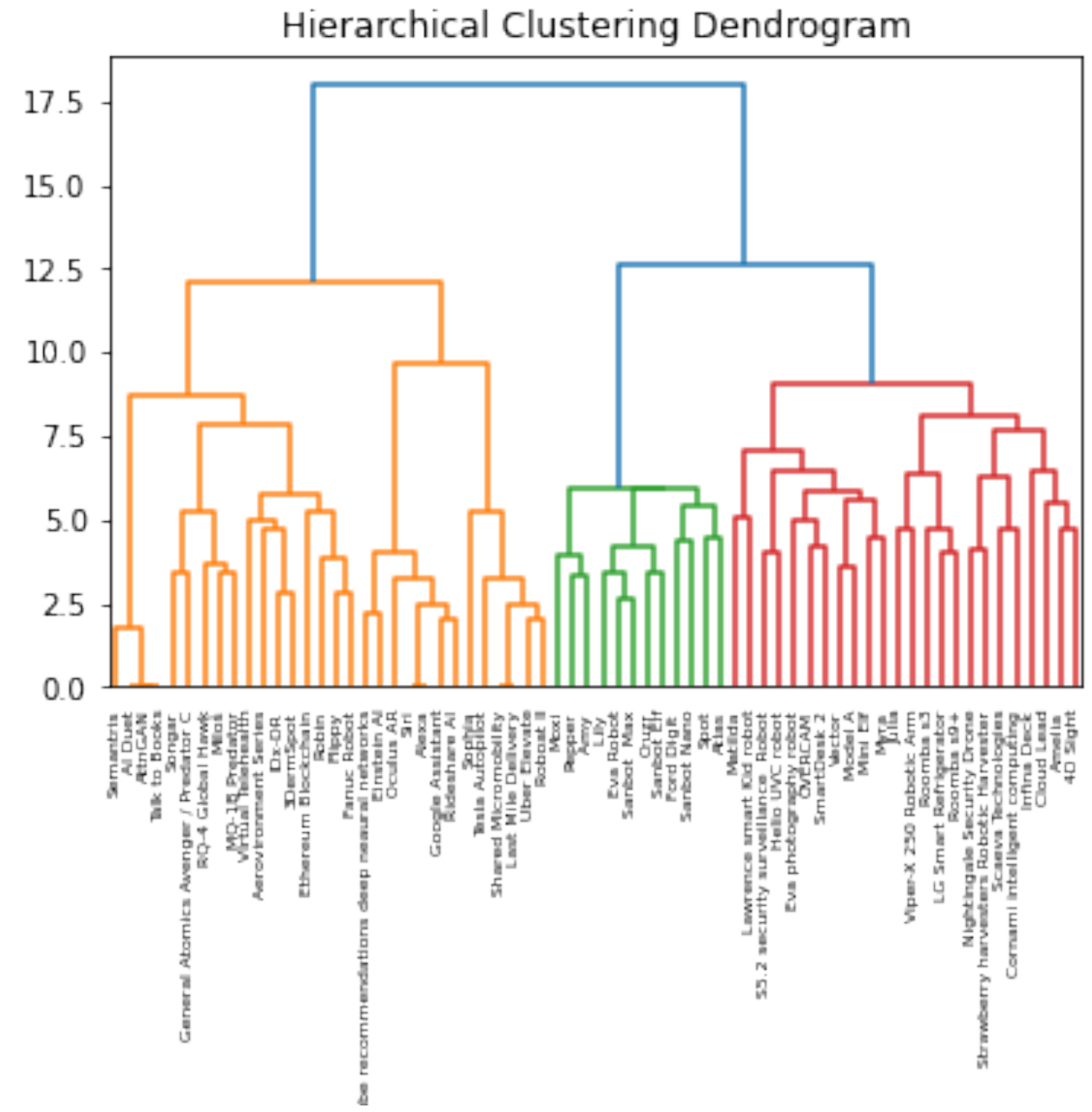

Fig. 4. Heriarchical Agglomerative Clustering Dendrogram

### 3) *Third Iteration*

And finally, the third iteration of the tool is a web app running on the Django web framework. The web app offers multiple advantages over the previous tools while maintaining any advantages gained along the way.

The biggest change on the back end allows the data to be saved directly into a database, whereas both of the previous solutions required exporting the data from one place to be imported in another for analysis.

The second big change is the integration of Python to the logic and capabilities pipeline, enabling the use of SciKitLearn and many other Python libraries for machine learning capabilities.

# V. RESULTS

## A. *AGGLOMERATIVE CLUSTERING*

We used the Python library, SciKitLearn, to implement modern machine learning techniques like agglomerative clustering to suggest appropriate clusters and hierarchies to begin to assemble a comprehensive taxonomy of autonomous entities.

The validation of the POC comes in the form of the system's ability to generate a complete taxonomy of every entity entered.

We employed an agglomerative clustering method based on the Ward algorithm[13]. This algorithm minimizes the sum of squared differences within all clusters. Unlike a k-means[14] type approach, the Ward algorithm uses a hierarchical approach to clustering while still trying to minimize variance.

This approach becomes very computationally expensive on large sample sizes because the algorithm checks every possible connection at every level. One method to reduce the computational requirements is to implement a connectivity matrix with connectivity constraints. Because our sample size was relatively small and the number of clusters relatively small as well, and because we did not want to introduce any bias in terms of the weights, we did not use any connectivity constraints.

The output of clustering was then used to create a dendrogram chart (Fig. 4). Again, while this approach for classification works well for relatively small sample sizes, a dendrogram like the one shown becomes impractical with larger taxonomies.

## B. *INTERNAL AND EXTERNAL VALIDITY*

We start with a brief examination of how we approached internal and external validity with regard to a new model. Internal validity asks if the model was designed, built, and operated correctly, and deals with a specific case. External validity asks if the results are valid and generalizable.

### 1) *Internal Validity*

- **Internal Validity** – Was this specific model designed, built, and operated correctly?

Our primary means of internal validation is the detailed information from the survey. We have already presented the methodology and process behind the survey used to gauge the dimensions and classes in the Riskyishness construct. Our statistical analysis demonstrates the relationship of how entities are treated by the Riskyishness construct, and the outcome of human-relatable clusters from the agglomerative clustering.

### 2) *External Validity*

- **External Validity** – Are the results valid and generalizable?

Given the short time frame of this pilot (one semester), we were unable to solicit as much feedback of the final product as we would have liked. However, we believe that the statistical validation provided demonstrates the generalizability with a sample size of n=86 different types of autonomous entities. The use of the clustering methodology is well established within the academic community.

## C. *DSRM FRAMEWORK EVALUATION METHODS*

- To address **Observational** concerns we have created an artifact to allow the user to enter any entity and receive a riskyishness score. In initial spot testing, we have found that score to be meaningful in both relative and absolute terms.

- To address **Analytical** concerns, we have completed the third iteration of an artifact to provide an interface allowing users to add an entity. We expect to collect 100+ samples (n) of a wide variety of Autonomous Entities to provide external validation of the method's generalizability.

- We used machine learning and statistical methods from Python, SciKitLearn, R Studio, and SAS Enterprise Miner to provide traditional statistical metrics.
- Further, the use of the clustering to categorize the entities demonstrates that the dimensions provide a meaningful scale from which to measure disparate dimensions.
- Entities like Siri and Alexa end up in the same branch, and factory robots in another, and military in another, and consumer robots in another, demonstrating that the math within the clustering, and based on the Riskyishness dimensions, leads to a human-interpretable and logical taxonomy.
- To demonstrate **Experimental** validity, we sent a qualitative survey to CGU scholars and peers. We created an initial draft of this instrument, received feedback from a small test group, and then created a second iteration that has been publicly shared. With the feedback obtained, we can fine-tune the user-facing artifact, as well as demonstrate internal validity.
- To address **Testing** concerns, we demonstrate the use and application by creating a webapp (3rd Iteration of the Tool) built on the Django framework. Django is Python-based and manages the database connections to separate the data from the application.
- **Descriptive** evidence has been provided by an exploratory literature review, as well as statistical support from the first survey. In line with DSR evaluation, key descriptions will include: functionality, completeness, consistency, accuracy, performance, reliability, usability, and fit with the 'riskyishness framework.'

## VI. Discussion and Conclusion

### A. Discussion of Limitations

We have demonstrated the application and use of DSRM to create a model for understanding Autonomous Entities, however a key tenet of DSR is the feedback loop from the end users to validate the final product. Unfortunately, due to the one-semester timeframe of the project, we have not been able to collect as much external validation as we would have liked.

With regards to statistical limitations, we recognize that much of the study was conducted as exploratory, and we sacrificed some statistical rigor in the interest of collecting sufficient representative data to test the methodology and artifacts.

In the future, greater attention can be paid to creating an experiment that includes double-blind evaluation of the methods and artifacts. Both can be measured using standard quantitative investigative techniques. Unfortunately, like much real-world research, the limitations lie not in the researchers' ability or desire, but in the financial and time constraints.

### B. Future Research

We imagine an open-source web project to add entities to a database/library to create a comprehensive taxonomy of the world's autonomous entities.

With regards to the methodology, we anticipate future research in multiple areas. A primary area of focus should be the refinement of the techniques required to articulate classes. We did not have time to run a regression analysis between respondents' views on the descriptions and their impression of the necessity of the dimension or class.

Future research will also focus on categorizing the different entities into buckets by determining common features that define a particular bucket and the hierarchical relationships that exist between the different classes of entities.

With regard to the specific algorithm used for clustering, while this worked very well for the small sample sizes, it would not scale to a global taxonomy as is. Therefore, appropriate directed research looking at the specifics of the clustering would add robustness to the solution.

Further general investigation could also focus around defining an OLS or PLS-SEM[15] type model to explain the relationship of user to Riskyishness interpretation or application.

One could also apply a UTAUT[16] type approach to acceptance of the methodology and tool, with regards to user acceptance of Riskyishness as a construct.

And finally, as a hypothetical construct, the combinatorial term "Riskyishness" needs to be independently operationalized so that it can be measured quantitatively and qualitatively to prove generalizability beyond the scope of autonomous entities.

### C. Conclusion

We have demonstrated the methodology for creating a construct that can be used to evaluate entities, as well as facilitating conversations across academic disciplines and corporate hierarchies. By refining the terms used in the scales, it allows for easy communication of some very complex ideas.

In one conversation with a participant, she pointed out that the term "UI/UX" might not be familiar to many people. She had a background in finance and had been CFO of multiple global organizations and Executive Director of the CFTC and was not familiar with the term. One researcher pointed out that although she had not been familiar when starting the process, it took about "2 minutes" for her to learn what UI/UX means, and now she had an industry term in her toolbox if she needs to talk to designers or anyone else.

It was this type of transdisciplinary communication that we had hoped to facilitate, and we believe the classes, dimensions, and scales were very successful from this perspective. We have found that discussions both internally and externally are guided by the framework towards a logical conclusion – often leading to an actionable result.

In conclusion, we found the pilot to be highly beneficial towards articulating a framework that benefits not only the academic community, but also real-world users seeking to understand the complex challenges associated with modern autonomous entities and technology. We believe we have addressed both philosophical and application concerns and maintained a high level of rigor to demonstrate DSRM at the crucial yet rapidly blurring intersection of humans and technology.

## Appendix A - Lexicon of Terminology Used in the Pinocchio Taxonomy of Autonomous Entities

**Dimension** - A specific characteristic or defining feature. Each dimension should be unique and distinct from any other dimension. Our Taxonomy has 25 dimensions in 6 classes.

**Class** - A group of dimensions. Each class should be unique and distinct from any other class. The groupings in v.0.0.3 are statistically arbitrary, defined by the researchers, and based on domain expertise.

1. **Physical Class** - Features or abilities that describe the entity's physical relationship to the world and its capacity to physically affect change.
    1. **Size** - The physical dimensions of the Autonomous Entity.
    2. **Locomotion** - The movement capabilities of the entity.
    3. **Manipulation** - Manipulation refers to an entity's control of its environment through selective contact.
    4. **Adoption** - How many, and/or how widespread is the entity.
    5. **Weaponry** - Any part of the entity that can be used to cause physical harm in either a defensive or offensive capacity.
2. **UI/UX Class** - Features or abilities that describe the UI/UX interaction with humans.
    1. **Human Interaction** - The extent to which the entity's interface bidirectionally engages a human being.
    2. **# of Human Senses addressed by UI/UX** - Number of human senses that are addressed by the UI/UX.
3. **Ethical / Philosophical Class** - Philosophical and ethical considerations of the entity.
    1. **Local Comprehension** - The range of technology and sensors on the entity that allow it to sense its local environment.
    2. **Global Comprehension** - The range of technology that allows the entity to understand a global environment.
    3. **Decision Making** - Ability of the entity to understand the meaning of a specific situation based on the inputs (local and global) and to respond.
    4. **Contextualization -** Ability of the entity to understand the specific situation with regard to the generalized situation based on the specific inputs (local and global).

5. **Self-Preservation** - The ability of the autonomous entity to protect itself from harm and destruction.
    6. **Complexity -** The extent to which an entity is linked to other entities by structural or existential dependencies.[17]
4. Application Class - Characteristics that describe the designer's intent and capability when designing the entity.
    1. **Intended Use** - The designer's intended use for the entity.
    2. **Financial Resources** - The long-term financial stability over the course of the entity's entire lifecycle.
5. Security / Privacy Class - Those characteristics or features that affect the security of the entity and the privacy of any information to which it has access.
    1. **Susceptible to Outside Influence** - How susceptible is the entity to outside influence.
    2. **3rd Party Dependencies** - The extent to which the entity is dependent on existing technology and its testing and adoption.
    3. **Data Collection** - The extent, type, and sensitivity of data collected.
    4. **Data Storage** - How, where, and for how long is the data stored?
    5. **Data Usage** - How is any collected data intended to be used?
6. Embodiment Class - Those characteristics or features that may approximate human emotion.
    1. **Physicality** - How aware the entity is of its physical presence.
    2. **Experiential** - How capable is the entity of practical contact and gaining awareness.
    3. **Emotional** - How capable is the entity of displaying and responding to human emotions.
    4. **Social -** How capable is the entity of representing or expressing itself in social settings and organized communities.
    5. **Transcendental** - Is the entity capable of understanding and expressing itself in spiritual and nonphysical realms.

# APPENDIX B - LEXICON OF TERMINOLOGY USED IN THE PINOCCHIO TAXONOMY OF AUTONOMOUS

| | | 0 | 1 | 2 | 3 | 4 |
|---|---|---|---|---|---|---|
| **Class** | **Dimension** | **Lowest** | **Associated Riskyishness Increase ——>** | | | **Highest** |
| Physical | Size | No Physical Dimension | Hand Held or Smaller | Human, Dog, Refrigerator | Car or Cessna | 18 Wheeler, Tank, or larger |
| | Locomotion | Immobile | Carried | Limited Mobility | Complete Mobility in one environment | Complete Mobility |
| | Manipulation | None | Indirect (Magnetic, Air, Etc...) | Limited and Incapable of harm | Semi-Articulated | Articulated |
| | Adoption | Iterative Prototypes / Hobbyists / Students | Niche Use Commercial / Industrial / Entertainment | Consumer Adoption | Widespread Consumer / Commercial / Industrial Adoption | Widespread use in Infrastructure or Military |
| | Weaponry | None | Kamikaze | Pseudo Weaponry | Significant Industrial Capability or Light Weaponry | Significant Offensive Weaponry or Nuclear Industrial capability |
| UI/UX | Human Interaction | None | Terminal Only | Some UI | Extensive UI but non-immersive (some Haptic feedback) | Immersive |
| | # of Human Senses addressed by UI/UX | 0 | 1-2 | 3 | 4 | 5 |
| Ethical / Philosophical | Local Comprehension | None | Analog | Sensors | External Data in a Sandboxed Environment | External Data via Internet |
| | Global Comprehension | None | Wi-Fi / LAN Only | Sandboxed | Limited On-Board AI - no control of primary systems | On-Board AI - complete control of primary systems |
| | Decision Making | None | Analog | Semi-Autonomous in a constrained environment | Semi-Autonomous with human review in an unconstrained environment | Autonomous |
| | Contextualization | None | Analog | Sandboxed | Intra-Net (Big Data - Boeing Digital Twin) | Internet (Big Data) |
| | Self Preservation | None | Passive Defense Only (Stealth, speed, flight, etc.) or Analog sensor hazard avoidance | Intelligent Avoidance | Active Defense - ECM, Flares, Smoke, Etc. | Offensive Defense - Ability to eliminate threats either reactively or proactively |
| | Complexity | No AI | On-Board Analog AI | Sandboxed Digital - some reliance on AI training | Limited On-Board AI Un-sandboxed, significant reliance on training data | Internet-Based AI, complete reliance on training data |
| Application | Intended Use | Consumer Recreational | Consumer Office / Medical | Consumer Mobile/ Non-industrial Commercial / Big Medicine or Pharma | Heavy Industrial / Large Commercial | Military |
| | Financial Resources | Long-term stable high capital / high capital to upkeep ratio | Short-term stable high capital / high capital to upkeep ratio | Low initial Capital investment/ high ratio | Medium Initial Capital Investment / Medium Ratio | Fluctuating Capital to upkeep requirement ratio |
| Security / Privacy | Susceptible to Outside Influence | None | External Damage via Physical Tampering | Some influence of sensors but no change in behavior | Susceptible to in-person, on-site attacks | Susceptible to remote attack |
| | 3rd Party Dependencies | None | Some dependency on well-tested technologies | Significant Dependency on well-tested technologies | Some dependency on new or untested technologies | Significant dependency on New or untested technologies |
| | Data Collection | None | Some Anonymized Data | Significant Anonymized Data that can be de-anonymized | Some non-sensitive individual data (video game trophies, Netflix queue, application preferences) | Sensitive personal information (medical, passwords, financial, military) |
| | Data Storage | None | Local and Temporary | Local and Permanent | Remote and Temporary | Remote and Permanent |
| | Data Usage | None | Strictly for internal use | Any sharing for non-profit and non-political reasons | Any sharing for profit, politics, law enforcement, etc... | Any sharing that is constitutionally prohibited |
| Embodiment | Physicality | None | Has Euclidean understanding of its physical form | Has 4D understanding (Time) | Can change physical embodiment | Can proactively evolve its form and replicate |
| | Experiential | None | Grades K,1-3 | Grade school | Highschool | College + |
| | Emotional | None | Recognize and feel basic emotions: fear, anger, sadness, joy | Emotions guided by self-preservation | Generosity, Philanthropy, Greed, Jealousy | Postulate - make complex emotional decisions |
| | Social | None | Awareness of the other | Interact with the other | Transact with the other | Govern, ability to create and control policies for itself and others |
| | Transcendental | None | Has minimal understanding of being Self-Aware | Has some understanding of being a part of a bigger entity like Universe | Has a basic understanding of complex consciousness like concepts of Infinity and Supernatural | Has a complex understanding of Transcendental Concepts. ex: Nothingness, Negative Infinity |